# Low Precision RNNs: Quantizing RNNs Without Losing Accuracy

Supriya Kapur, Asit Mishra, Debbie Marr

Accelerator Architecture Lab, Intel Corporation
kapursu@oregonstate.edu, asit.k.mishra@intel.com, debbie.marr@intel.com

## ABSTRACT

Similar to convolution neural networks, recurrent neural networks (RNNs) typically suffer from over-parameterization. Quantizing bit-widths of weights and activations results in runtime efficiency on hardware, yet it often comes at the cost of reduced accuracy. This paper proposes a quantization approach that increases model size with bit-width reduction. This approach will allow networks to perform at their baseline accuracy while still maintaining the benefits of reduced precision and overall model size reduction.

## 1. INTRODUCTION

Recurrent Neural Networks (RNNs) are a type of neural network that have a memory aspect, allowing the network to draw on its previous knowledge. In this way, RNNs have a loop of cyclical information that can be used to interpret and assess the current information being processed. RNNs are typically used for translation and language modeling, and have become increasingly popular both in industry and research due to their impressive performance [3,7,8,10]. Owing to their state of the art accuracy, these networks have drawn considerable interest from the hardware community and various studies have proposed methods to reduce the runtime complexity of RNNs using quantization schemes [4,6]. Lowering the precision of data types used in these networks lowers the storage size, runtime memory requirements, and makes RNNs amenable to be deployed on low-memory hardware platforms.

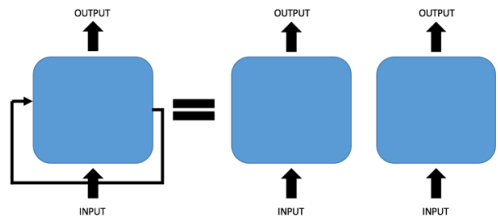

**Figure 1.** *An RNN cell unrolled.*

Popular methods to quantize a network include reducing bit-widths in order to shrink both the size of the network by directly reducing the number of bits needed for weights, biases, and activations, while also decreasing the complexity of the computations. Bit-width reduction does however come at the cost of network accuracy, as drastic loss in accuracy can be seen when networks are reduced from their original 32bit width.

This paper investigates the quantization of RNNs without sacrificing network accuracy, with the overall goal of efficient hardware implementation. Our scheme will pair reduction in bit width with increase in the model dimension, i.e. increasing the number of neurons in a layer. Our method can retain accuracy while still reducing the network size and runtime memory usage, thus achieving baseline accuracy at a fraction of the memory cost of the baseline network.

We use a quantization method very similar to that proposed in [6]. We study the quantization scheme along with increasing the number of neurons in a layer extensively on a language model on the Penn Treebank (PTB) dataset [2]. We then apply our scheme of increasing the number of neurons and lowering the bit-width on Baidu's Deep Speech model [5]. Results show that the Deep

Speech model can be quantized down to 33% of its original size, while still maintain baseline accuracy, thus proving the efficacy of our approach.

## 2. NEURON INCREASE APPROACH

The number of neurons in each layer of a neural network is representative of the computing complexity of the network. Typically, the number of neurons directly affects the accuracy of the network. Increasing the number of neurons in each layer will increase network accuracy, as it will allow for more discriminative power within each layer, as well as from layer to layer.

Our neuron increase approach will target certain layers in an RNN model in order to re-gain accuracy that was lost with bit-width quantization. This will be done by first evaluating the network, layer by layer, in order to identify smaller layers that will benefit from an increase in neurons while still maintaining the reduced model size. Once the layers with neuron increase potential are identified, they will be altered by a factor of a set percentage in order to increase neurons. All other layers will retain their original amount of neurons.

Increasing or decreasing the neurons in a layer has a linear relationship in terms of the number of compute operations with the layer size. For this reason, the layers targeted for neuron increase are the never the network's larger layers. We find that increasing the neurons in certain layers combined with bit-width quantization effectively retains accuracy. Model size is slightly increased once neurons are added to certain layers, but values, such as weights, biases, and activations, are still stored in limited bit-widths. As a result, the network will retain the benefit of smaller storage size and less complex computations. Computations can be simpler if instead of floating point compute one operates in integer mode.

The neuron increase approach was studied extensively through a language model on the PTB dataset and then applied to the Deep Speech model, achieving the goal of a quantized network which retains accuracy on both models.

## 3. STUDIES ON A LANGUAGE MODEL USING PTB

In order to evaluate the rate in which accuracy is deteriorated in RNNs when the bit-width is reduced, a language model was studied. Using the quantization methods discussed in [6], both the weights and activations of a Long-Short-Term Memory (LSTM) RNN model were reduced. The neuron increase approach was then applied in order to see the effect it can have on re-gaining the lost accuracy in quantized RNNs. The model used for testing consisted of one LSTM layer containing 300 neurons, and was tested on the PTB dataset. This dataset, considered small in size, contains 10000 unique words. This quantized RNN model, called Bit-RNN in [6], measures accuracy in perplexity per word (PPW). Table 1 summarizes the results of our quantized experiments on this model.

**Table 1.** *Bit-RNN model PPW with varying bit-widths.*

| Weight Bit-Width | Activation Bit-Width | | | | |
|---|---|---|---|---|---|
| | *32* | *16* | *8* | *4* | *2* |
| *32* | 108.7 | 108.8 | 108.3 | 109.2 | 111.4 |
| *16* | 112.6 | 112.7 | 112.4 | 113.3 | 114.8 |
| *8* | 113.3 | 112.8 | 112.4 | 112.1 | 115.1 |
| *4* | 114.3 | 114.5 | 114.8 | 115.2 | 116.7 |
| *2* | 141.1 | 138.6 | 139.1 | 138.5 | 152.2 |

Table 1 shows that decreasing the bit-width of the LSTM model has a much more drastic effect on the model accuracy, as accuracy only drops by approximately 1.0% when the activation bits are halved, but drops by approximately 8.2% when the weight bits are halved.

Using the quantized Bit-RNN models, the neuron increase approach was tested to see the relationship between increasing neurons and network accuracy. Because the Bit-RNN model consists of only one layer, the neuron increase approach was applied without needing to do layer by layer network evaluation. Once the Bit-RNN model was quantized and the neuron increase approach was applied, the model size reduction and runtime memory usage was evaluated. The quantization methods used leads to a model size reduction of $k/32$, where *k* represents the specified weight bit, and reduces the runtime memory usage by $k/32$ as well, where *k* represents the specified activation bit. As stated earlier, increasing the number of neurons will have a linear effect on model size. However, with model size increase, we also lower the bit-width which lowers the overall memory requirements.

**Table 2**. *Bit-RNN model PPW with varying bit-widths and neurons.*

| Model Configuration (Weight x Activation) | Number of Neurons | | |
|---|---|---|---|
| | *300* | *450* | *1000* |
| *4 x 4* | 115.2 | 108.8 | 108.4 |
| *4 x 2* | 116.7 | 111.7 | 113.1 |
| *2 x 4* | 138.5 | 130.6 | 128.4 |
| *2 x2* | 152.2 | 148.2 | 148.0 |

Table 2 shows the effect of increasing the number of neurons in the network by 1.5x and ~3x. Once the neurons are increased by 1.5x, full accuracy is regained with 4-bit weights and 4-bit activations. Accuracy is within 2% with 4-bit weights and 2-bit activations when number of neurons are increased by 1.5x. We find the accuracy to plateau once neurons are increased beyond 1.5x. Additionally, the 4-bit weights and 4-bit activations model is only 13% in size of the baseline model size and runtime memory usage, and when the neurons are increased by 1.5x of their original amount, the network can surpass baseline accuracy while only having a 6% jump in model size and runtime memory usage.

## 4. DEEP SPEECH MODEL STUDIES

The Deep Speech model is a speech recognition system that can perform well on large datasets. This model serves as a much larger arena to prove the effectiveness of bit-width reduction when paired with neuron increase. The Deep Speech model used in the following experiments is constructed with six layers; five fully connected (FC) layers and one recurrent bidirectional [9] LSTM layer. Three of the five FC layers consist of 2048 neurons, the FC layer that feeds into the LSTM layer contains double that, as does the entirety of the LSTM layer (2048 neurons in the backward and in the forward cell), and the final layer of the network contains the same number of neurons as characters in the language being used.

The Deep Speech model was tested by first quantizing the network and then applying the neuron increase approach. The models were trained using the LibriSpeech training dataset [11], which contains approximately 1000 hours of speech, and the TED-LIUM training dataset [1], which contains approximately 200 hours of speech. The models were tested on the TED-LIUM testing dataset. Accuracy is measured using the word error rate (WER).

The Deep Speech model was quantized in two different forms in order to see the effect of quantization on the accuracy of the FC layers as compared to quantization in only the recurrent LSTM layer. This was done by first quantizing only the FC layers via their weights and biases, and in a separate model only quantizing the recurrent LSTM layer. Due to resource limitations on a GPU, the recurrent LSTM layer was only quantized by its activations, not weights. Once quantized, the respective network models were evaluated for neuron increase potential. The first quantized model had a neuron increase in the first, second, and fifth layers of the model, as these layers are relatively smaller layers in the network, and thus will not greatly effect model size when their number of neurons are increased. The second quantized model did not receive any neuron increases, as data from the previous section's study suggests that when only activations are quantized, model accuracy is maintained.

**Table 3.** *Deep Speech model WER with reduced FC layers and increased neurons.*

| Configuration | Δ WER |
|---|---|
| 8 Bit FC Layer | 3.3% |
| 8 Bit FC Layer + 25% Neuron Increase | 2.1% |
| 8 Bit FC Layers + 50% Neuron Increase | -1.7% |
| 4 Bit FC Layer | 5.9% |
| 4 Bit FC Layer + 25% Neuron Increase | 3.4% |

Table 3 shows the difference in WER from the baseline model accuracy, and reveals that the accuracy increases when neurons are added to the FC layers at a 50% increase and a 25% increase. The results show that even the 4-bit FC layer model is able to retain accuracy when the neurons in the first, second, and fifth, layer are increased by 50%. The effect of the added neurons on model size can also be measured. In the 4-bit FC layer model, once neurons are increased by 25% of their original size in layers one, two, and five, the model size is only increased by 0.6%, and when the neurons are increased to 50% of their original size, baseline accuracy is retained and the model size only jumps 1.2%.

**Table 4.** *Deep Speech model WER with reduced recurrent layer.*

| Configuration | Δ WER |
|---|---|
| 8 Bit Recurrent Layer Activations | 0.4% |
| 4 Bit Recurrent Layer Activations | 2.1% |

Table 4 shows that decreasing the activations in the recurrent layer has a minimal effect on accuracy, even when reduced to 4 bits. Additionally, runtime memory usage is reduced to 78.6% of its original size when the LSTM activation bits are quantized to 8 bits, and only 71.5% of its original size when the LSTM activation bits are quantized to 4 bits.

## 5. CONCLUSIONS

Results from a language model on the PTB dataset and the Deep Speech model study both conclude that bit-width quantization paired with neuron increases can effectively retain accuracy. The language model on the PTB dataset study revealed that quantizing the weights has a much larger effect on model accuracy when compared to quantizing activations, and that the neuron increase approach can re-build accuracy with only a 50% neuron increase. The Deep Speech model study proved that the model can retain baseline accuracy when all FC layers are quantized to 4 bits and only three out of the six layers have a neuron increase of 50%. Additionally, model size is only increased by 1.2% in order for accuracy to be retained in the 4-bit model.

Our future work on this topic will include quantizing the weight bits of the recurrent layer in combination with the activation bits and quantizing the recurrent layer and FC layers simultaneously.

## 6. REFERENCES

[1] A. Rousseau, P. Deléglise, and Y. Estève, "Enhancing the TED-LIUM Corpus with Selected Data for Language Modeling and More TED Talks", in Proceedings of the Ninth International Conference on Language Resources and Evaluation (LREC'14), May 2014.

[2] Ann Taylor, Mitchell Marcus, and Beatrice Santorini. The penn treebank: an overview. In Treebanks, pp. 5–22. Springer, 2003.

[3] Donahue, Jeffrey, et al. "Long-term recurrent convolutional networks for visual recognition and description." *Proceedings of the IEEE conference on computer vision and pattern recognition*. 2015.

[4] Guan, Tianchan, Xiaoyang Zeng, and Mingoo Seok. "Recursive Binary Neural Network Learning Model with 2.28 b/Weight Storage Requirement." *arXiv preprint arXiv:1709.05306* (2017).

[5] Hannun, Awni, et al. "Deep speech: Scaling up end-to-end speech recognition." *arXiv preprint arXiv:1412.5567* (2014).

[6] He, Qinyao, et al. "Effective Quantization Methods for Recurrent Neural Networks." *arXiv preprint arXiv:1611.10176* (2016).

[7] J. Lee, K. Kim, T. Shabestary and H. G. Kang, "Deep bi-directional long short-term memory based speech enhancement for wind noise reduction," *2017 Hands-free Speech Communications and Microphone Arrays (HSCMA)*, San Francisco, CA, 2017, pp. 41-45.

[8] LeCun, Yann, Yoshua Bengio, and Geoffrey Hinton. "Deep learning." *Nature* 521.7553 (2015): 436-444.

[9] M. Schuster and K. K. Paliwal, "Bidirectional recurrent neural networks," in *IEEE Transactions on Signal Processing*, vol. 45, no. 11, pp.2673-2681, Nov 1997.

[10] Miller, Clifford B., and C. Lee Giles. "Experimental comparison of the effect of order in recurrent neural networks." *International Journal of Pattern Recognition and Artificial Intelligence* 7.04 (1993): 849-872.

[11] Panayotov, Vassil, Chen, Guoguo, Povey, Daniel, and Khudanpur, Sanjeev. Librispeech: an asr corpus based on public domain audio books. In ICASSP, 2015.